\documentclass{bmvc2k}

\usepackage{amsmath}
\usepackage{amssymb}
\usepackage{todonotes}
\usepackage{comment}
\usepackage{bm}

\title{3D-GMNet: Single-View 3D Shape Recovery as A Gaussian Mixture}

\makeatletter
\def\addauthorX#1#2#3{
  \advance\bmv@nauthors1
  \bmv@savenewbox{authname\the\bmv@nauthors}{\bmv@RenderAuthorName{#1}}
  \bmv@savenewbox{authnameFN\the\bmv@nauthors}{\bmv@RenderAuthorNameFN{#1}{#3,2}}
  \bmv@savenewbox{authmail\the\bmv@nauthors}{\bmv@RenderAuthorMail{#2}}
  \bmv@newcount{authinst\the\bmv@nauthors}
  \bmv@setcount{authinst\the\bmv@nauthors}{#3}
  \bmv@edefappend{\bmv@auths}{\the\bmv@nauthors}
}
\makeatother
\addauthor{Kohei Yamashita}{kyamashita@vision.ist.i.kyoto-u.ac.jp}{1}
\addauthorX{Shohei Nobuhara}{nob@i.kyoto-u.ac.jp}{1}
\addauthor{Ko Nishino}{kon@i.kyoto-u.ac.jp}{1}

\addinstitution{
 Kyoto University\\
 Kyoto, Japan
}
\addinstitution{
JST, PRESTO\\
Saitama, Japan
}

\runninghead{K. YAMASHITA, S. NOBUHARA, K. NISHINO}{3D-GMNet}

\def\onedot{\bmvaOneDot}

\def\ie{\emph{i.e}\onedot}

\def\wrt{w.r.t\onedot} 
\def\etal{\emph{et al}\bmvaOneDot}

\begin{document}

\maketitle

\begin{abstract}
In this paper, we introduce 3D-GMNet, a deep neural network for 3D object shape reconstruction from a single image. As the name suggests, 3D-GMNet recovers 3D shape as a Gaussian mixture. In contrast to voxels, point clouds, or meshes, a Gaussian mixture representation provides an analytical expression with a small memory footprint while accurately representing the target 3D shape. At the same time, it offers a number of additional advantages including instant pose estimation and controllable level-of-detail reconstruction, while also enabling interpretation as a point cloud, volume, and a mesh model. We train 3D-GMNet end-to-end with single input images and corresponding 3D models by introducing two novel loss functions, a 3D Gaussian mixture loss and a 2D multi-view loss, which collectively enable accurate shape reconstruction as kernel density estimation. We thoroughly evaluate the effectiveness of 3D-GMNet with synthetic and real images of objects. The results show accurate reconstruction with a compact representation that also realizes novel applications of single-image 3D reconstruction.

\end{abstract}

\section{Introduction}
Single-view 3D shape recovery finds many applications in a wide range of domains including robotics, mixed reality, and graphics. Image-based 3D reconstruction is, however, a fundamentally ill-posed problem due to the inherent loss of dimensionality through image projection. Past methods have leveraged constraints arising from projective geometry~\cite{scharstein2002taxonomy,SfM-survey,agarwal2011building,schonberger2016structure}, radiometric surface properties~\cite{alldrin2008photometric,agrawal2006range,hernandez2007non}, and optical imaging properties~\cite{favaro2008shape,nayar1994shape} to arrive at unique 3D reconstructions. For the even more underconstrained single-view 3D shape recovery, recent works have shown that convolutional neural networks (ConvNet) can be trained end-to-end to impose effective priors for accurate reconstruction~\cite{Fan_2017_CVPR,Jiang_2018_ECCV,Wang_2018_ECCV,choy20163d,lin2018learning}.

Past methods on single-view 3D shape reconstruction have chiefly employed conventional representations of geometry: point clouds, volumes, and mesh models. Each of these representations have their pros and cons. Point clouds are simple enough to learn their mapping from single images~\cite{Fan_2017_CVPR, Jiang_2018_ECCV, lin2018learning}, but they lack topological (surface) information. Volumes are straightforward to train and infer with 3D ConvNets~\cite{choy20163d, marrnet}. Their resolution is, however, limited in practice due to the inherent cubic memory cost. Although, in contrast, meshes efficiently represent object surfaces~\cite{Fan_2017_CVPR, kato2018renderer}, the non-parametric 2D representation makes constraining and reconstructing occluded sides of general objects challenging.

\begin{figure}
    \centering
    \includegraphics[width=1\columnwidth]{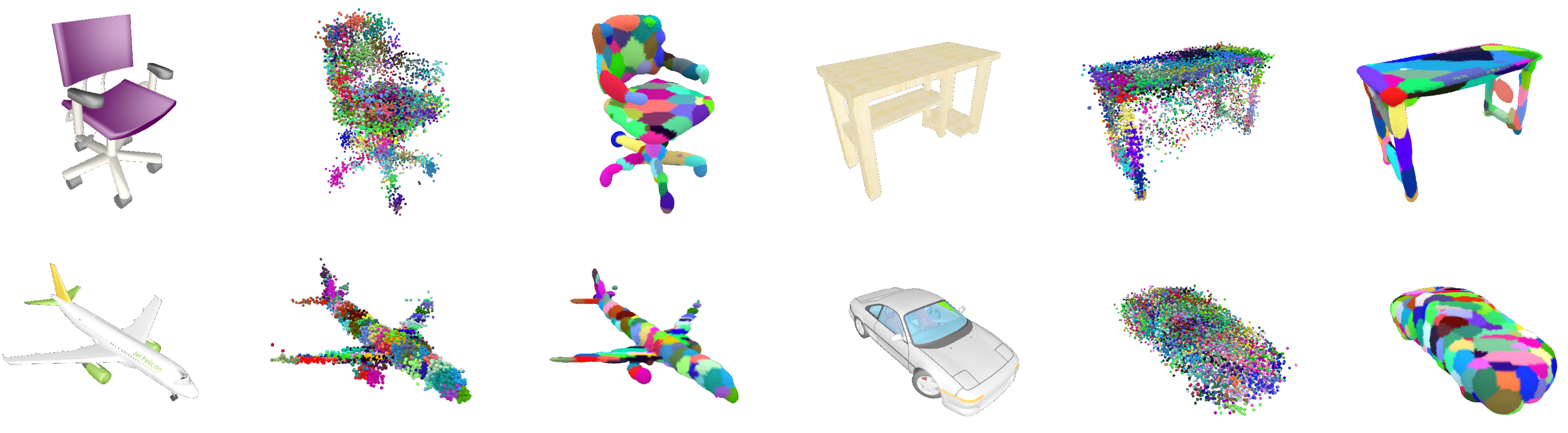}
    \caption{We introduce 3D Gaussian Mixture Network (3D-GMNet) for single-view 3D image-based reconstruction. From a single image, the network recovers object shape as a 3D Gaussian mixture which is extremely compact, integrates properties of conventional geometry presentations, and offers additional benefits for novel applications. In this figure, the object shapes are represented with 256 Gaussians.}
    \label{fig:gmm}
\end{figure}

In this paper, we derive a novel method for recovering 3D shape from a single image as a Gaussian mixture. As depicted in Fig.~\ref{fig:gmm}, our key contribution lies in enabling the reconstruction of the whole 3D geometry of an object from its single-view observation as a compact, analytical model that can be sampled as a point cloud, interpreted as a volume, and distilled into a surface. We train a ConvNet to learn this mapping from an image to a Gaussian mixture end-to-end by formulating 3D shape recovery as kernel density estimation. The 3D Gaussian mixture shape representation significantly reduces memory footprint compared to volume-based occupancy estimation approaches~\cite{choy20163d,marrnet} while providing a straightforward means for defining the surface unlike unstructured point cloud representations~\cite{Fan_2017_CVPR,Jiang_2018_ECCV}. Also in sharp contrast to mesh-based shape representations, the Gaussian mixture model enables the network to adaptively refine the shape topology.

Two recent works have demonstrated the advantages of representing 3D geometry as a Gaussian mixture, purely as a generator~\cite{hertz2020pointgmm} or from an input 3D mesh model~\cite{Genova_2019_ICCV}. Most closely related to our work, Genova \etal also demonstrated its use for single-view 3D shape recovery through network distillation~\cite{Genova_2019_ICCV}. Their representation is, however, inherently constrained to consist of axis-aligned 3D Gaussians, which fundamentally limits the ability to approximate general objects that can have angled structures. Our method is not limited to axis-aligned Gaussian mixtures, hence not bound to carefully axis-aligned objects. Furthermore, our 3D shape recovery is achieved in the viewer-centric coordinate system, \ie, the output shape is in the camera coordinate frame, which greatly expands the general applicability of the method, and also enables applications such as pose estimation.



We derive a deep neural network which we refer to as the 3D-GMNet that learns to output a set of parameters of a Gaussian mixture shape model that explains the input image and associated 3D model at training time. We propose two novel loss functions to train 3D-GMNet end-to-end. The first is the 3D Gaussian mixture loss, which evaluates the accuracy of the estimated Gaussian mixture shape model with regards to the target 3D shape. This is achieved by maximizing the likelihood of the Gaussian mixture which in turn is evaluated by considering the target 3D points as samples from the true distribution. The second is the 2D multi-view loss that evaluates the accuracy of the 2D projections of the Gaussian mixture to random viewpoints against the true silhouettes, \ie, the projections of the ground truth 3D shape to the same viewpoints. We show that these 3D and 2D losses work hand-in-hand to estimate accurate and effective Gaussian mixtures for general objects.

We conduct extensive experimental validation of the effectiveness of 3D-GMNet using images of both synthetic and real objects. We also demonstrate the advantages of the estimated 3D Gaussian mixture shape model over conventional geometry representations both qualitatively and quantitatively. Most important, we show that the reconstructed shape model is compact as it only requires the 3D mean and covariance for each mixture component. In addition, it admits a number of direct favorable applications, including controlled level-of-detail reconstruction via Gaussian mixture reduction, pose estimation, and distance measurement. The results show that 3D-GMNet achieves accurate single-image shape estimation with a representation that opens a new avenue of applications of image-based geometry reconstruction.



\section{Related Work}
\paragraph{Shape Representation}
Learning-based 3D shape estimation studies can be categorized by their shape representations: voxels~\cite{choy20163d,marrnet}, point clouds~\cite{Fan_2017_CVPR,Jiang_2018_ECCV,navaneet2019capnet}, patches~\cite{atlasnet}, mesh models~\cite{Wang_2018_ECCV, kato2018renderer}, primitive sets~\cite{niu_cvpr18,abstractionTulsiani17, Paschalidou2019CVPR, Genova_2019_ICCV,hertz2020pointgmm}, and learned functions~\cite{OccupancyNetworks, Park_2019_CVPR}. Wu \etal~\cite{marrnet} discretize the target 3D shape into a $128{\times}128{\times}128$ voxel grid and their neural network estimates the occupancy of each voxel. This is a memory-intensive approach, although it can handle 3D shapes of different topology in a unified manner. Lin \etal~\cite{lin2018learning} propose a network that estimates multi-view depth-maps from a single image. Groueix \etal~\cite{atlasnet} represents the target 3D shape by a collection of 3D patches. Although memory efficient, fusing multiple depth-maps or multiple patches into a single watertight 3D shape remains challenging. Mesh-based approaches~\cite{Wang_2018_ECCV, kato2018renderer} can make use of local connectivity of the 3D shape. Handling different topologies, however, becomes an inherently challenging task with meshes. Primitive-based approaches~\cite{niu_cvpr18,abstractionTulsiani17, Paschalidou2019CVPR} represent the target 3D shape as a collection of simple objects such as cuboids or superquadrics. They can realize a compact representation of the target volume, but cannot represent smooth and fine structures by definition. Mescheder \etal~\cite{OccupancyNetworks} train the network as a nonlinear function representing the occupancy probability of 3D object shape. Although highly scalable in resolution, it is a computation-intensive approach since the network should infer the probability for each and every sample. Saito \etal\cite{saito2019pifu} use Pixel-aligned Implicit Function to improve memory efficiency albeit specifically for human body shape recovery. Unlike these methods that recover a sampled volume of an implicit function,  we recover the parameters of an analytical implicit function, which is much more memory and computation efficient.

Our Gaussian mixture-based representation has the advantages of these conventional shape representations. It models not only the surface points but also the interior of the volume, with an efficient parameterization, \ie, a set of Gaussian parameters, and can generate a watertight 3D surface of arbitrary resolution as its isosurface. It can be considered as a probability density approach with Gaussian distributions, and also as a primitive-based approach with Gaussians as primitives. Additionally, in contrast to cuboid-based approaches, our shape representation can realize 3D registration with a simple canonical algorithm.

\vspace{-12pt}
\paragraph{Neural Networks for Mixture Density Estimation}
Mixture density network~\cite{Bishop94mixturedensity} is a method to predict a target multimodal distribution as a mixture density distribution. Bishop~\cite{Bishop94mixturedensity} introduced this network architecture with isotropic Gaussian basis functions. Williams~\cite{Williams_1996} extended it to utilize a general multivariate Gaussian distribution as the basis function. As described in Sec.~\ref{section:architecture}, our density estimation network is inspired by these works.

\section{3D Gaussian Mixture Network}

\begin{figure}
    \centering
    \includegraphics[width=\linewidth]{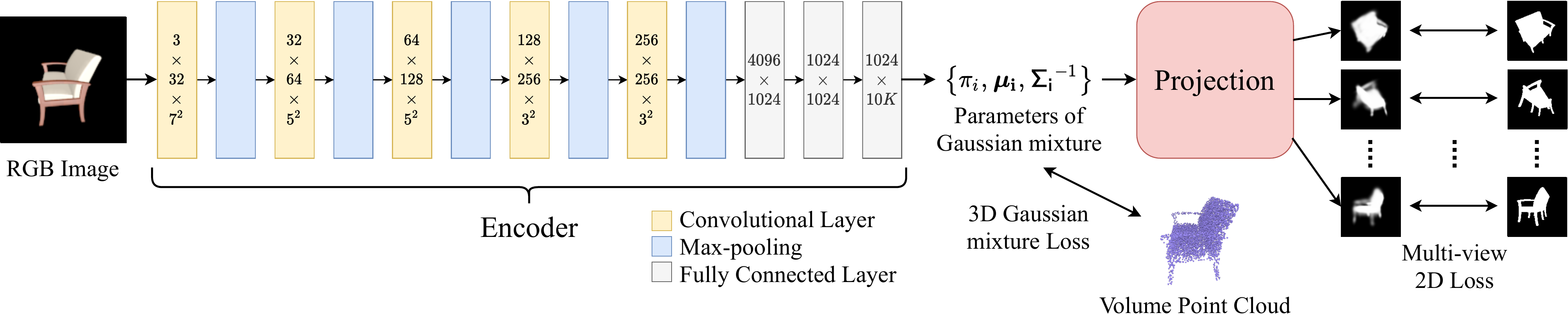}
    \caption{3D-GMNet is trained end-to-end by finding a Gaussian mixture that best represents the volume point cloud associated with the input image while also minimizing the discrepancy in their multi-view projections.}
    \label{fig:overview}
\end{figure}

Figure \ref{fig:overview} shows an overview of our 3D Gaussian Mixture Network (3D-GMNet).  Given a 2D image of an object, our 3D-GMNet estimates a set of parameters that defines a Gaussian mixture that best represents the 3D shape of the object in the input image. 


\subsection{3D Shape as A Gaussian Mixture}
\label{section:representation}

Our key idea is to consider the target 3D volume as a collection of observations of a random variable with a Gaussian mixture distribution. Suppose a ground-truth 3D shape is given as a volumetric 3D point cloud. We assume that this 3D point cloud samples the object volume, which can easily be computed from the 3D models from the training data. As such, we may regard them as voxels, too. Each one of the voxels is a sample of the random variable, and our goal when training the network is to estimate a 3D Gaussian mixture distribution that describes these samples best. 


A 3D Gaussian mixture distribution is defined as
\begin{equation}
    f_\mathrm{GM}(\bm{x}) = \sum_{i=1}^{K} \pi_i \phi(\bm{x}|\bm{\mu}_i,\mathsf{\Sigma}_i)\,,
    \label{eq:gm}
\end{equation}
where $K$ is the number of mixture components and $\{\pi_i\}$ are the mixing coefficients that sum to $1$, and each component $\phi(\bm{x}|\bm{\mu},\mathsf{\Sigma})$ is a 3D Gaussian with mean $\bm{\mu}$ and covariance $\mathsf{\Sigma}$. Note that the covariance matrix is not limited to be diagonal. Gaussian mixtures can approximate various kinds of distributions with an appropriate $K$. 3D-GMNet is trained to output the parameters of $f_\mathrm{GM}$ from a single input image.

Once the density function $f_\mathrm{GM}(\bm{x})$ is obtained, in addition to sampling a 3D point cloud or viewing it as a volume, we can extract the 3D object surface. Assume that we knew the volume of the object $V$, though it is not available in reality. The object surface is given as the isosurface of the density at $\tau = c/V$, where $c$ decides the level of thresholding. We approximate the unknown ${1}/{V}$ by the expectation of the density that can be computed analytically in closed-form
\begin{equation}
    \mathrm{E}\left[f_\mathrm{GM}(\bm{x})\right] = \int f_\mathrm{GM}(\bm{x})^2\mathrm{d}\bm{x} = \sum_{i=1}^K\sum_{j=1}^K\pi_i\pi_j\phi(\bm{x}=\bm{\mu}_i|\bm{\mu}_j,\mathsf{\Sigma}_i+\mathsf{\Sigma}_j)\,,
    \label{eq:expectationOfDensity}
\end{equation}
since $\mathrm{E}[f_\mathrm{GM}(\bm{x})] = \frac{1}{V}$ holds if $f_\mathrm{GM}(\bm{x})$ is identical to the true distribution $f(\bm{x})$. The parameter $c$ is determined experimentally in the evaluations in Sec.~\ref{sec:results}. By thresholding the target space with this value, we obtain the volumetric representation of the 3D shape, which can then be converted to a surface model using the marching cubes algorithm~\cite{lewiner2003efficient}.

\subsection{Network Architecture}
\label{section:architecture}

3D-GMNet outputs a set of parameters of a Gaussian mixture $\{\pi_i, \bm{\mu}_i, \mathsf{\Sigma}_i\}_i^K$. As depicted in Fig.~\ref{fig:overview}, the network has an encoder module to predict these parameters and a projection module to render multi-view 2D silhouettes. The encoder consists of 5 convolutional layers, 5 max pooling layers of kernel size 2, and 3 fully-connected layers. Each of the convolution layers is followed by a batch normalization layer and a leaky ReLU activation layer. Each fully-connected layer except the last one is followed also by a leaky ReLU layer.

After these layers, we introduce an output layer tailored for Gaussian mixture parameters to enforce constraints to make it a valid probability density function~\cite{Bishop94mixturedensity, Williams_1996}. The mean $\{\bm{\mu}_i\}$ should be a 3D position in Euclidean space $\bm{\mu}_i\in \mathbb{R}^3$, and we use an identity mapping for $\bm{\mu}_i$ as $\bm{\mu}_i = \bm{a}_{\bm{\mu}_i}$, where $\bm{a}_{\bm{\mu}_i}$ is the corresponding output of the last layer. To ensure that the coefficient $\{\pi_i\}$ sum to $1$, the output layer applies softmax activation.

The precision matrix $\mathsf{\Sigma}^{-1}_i$ of a Gaussian component should be a symmetric positive definite matrix, and can be decomposed as $\mathsf{\Sigma}^{-1}_i=LL^\top$ using Cholesky decomposition where $L$ is a lower triangular matrix. Thus our network predicts $L=\{l_{ij}\}$ instead of $\mathsf{\Sigma}_i^{-1}$ where
\begin{equation}
    l_{ij}=
    \begin{cases}
        a_{l_{ij}} & i>j\,,\\
        \exp(a_{l_{ij}}) & i=j\,,\\
        0 & \text{otherwise}\,,
    \end{cases}
\end{equation}
where $a_{l_{ij}}$ is the corresponding output of the last layer. Notice that this enforces $l_{jj}$ to be positive so that the mapping from $L$ to $\mathsf{\Sigma}^{-1}_i$ is bijective.

\subsection{3D Gaussian Mixture Loss}
\label{section:3dloss}
We provide 3D shape supervision at training time as voxels. To quantitatively evaluate the fit of the estimated Gaussian mixture to these voxels, we use the Kullback-Leibler (KL) divergence, which amounts to minimizing the cross entropy $-\int p(x)\log{q(x)}\mathrm{d}x$. By considering the target voxels as observations from the true distribution, we can compute this efficiently with Monte Carlo sampling
\begin{equation}
    L_\mathrm{3D} = -\frac{1}{|P|}\sum_{\bm{x}\in P} \log f_\mathrm{GM}(\bm{x})\,,
    \label{eq:loss3d}
\end{equation}
where $f_\mathrm{GM}(\bm{x})$ is the output of our network, $\bm{x} \in P$ is a sample from the target density $f(\bm{x})$ and $|P|$ is the number of sampled points. For training, we randomly sample a fixed number of 3D voxels from the original target voxels for each mini batch.

We also introduce a loss that encourages Gaussian components to be distributed within a distance $T$ from the object center
\begin{equation}
    L_\mathrm{dist} = \frac{1}{K}\sum_{i=1}^{K}\left\{\mathrm{ReLU}(|\bm{\mu}_i| - T)\right\}^2\,.
\end{equation}
In our experiments, we use $T=0.85$ to cover the entire object space.

\subsection{2D Multi-view Loss}\label{section:2dloss}

The 3D loss is not sufficient to recover accurate geometric shape, especially for general objects that can have thin, angled structures. For this, we also leverage 2D projections of the 3D shape and derive a differentiable 2D multi-view loss that evaluates the consistency of object silhouettes.


To generate a silhouette of a 3D Gaussian mixture of Eq\onedot~\eqref{eq:gm}, we use para-perspective projection\cite{Hartley:2003:MVG:861369} for each mixture component since it projects a 3D Gaussian as a 2D Gaussian. As a result, we obtain a 2D Gaussian mixture as a projection of our 3D Gaussian mixture shape representation. Note that perspective projection does not result in a Gaussian due to its nonlinearity. 
%
We can derive the para-perspective projection of a 3D Gaussian mixture (see supplementary material for details)
\begin{equation}
    d(\bm{x}) = \sum_{i=1}^{K}\pi_i \phi_{2D}(\bm{x}|\bm{\mu}'_i, \Sigma'_i)\,,
    \label{eq:gmm2d}
\end{equation}
where $\phi_{2D}(\cdot)$ denotes a 2D Gaussian of the form
\begin{equation}
    \phi_{2D}(\bm{x}|\bm{\mu}',\Sigma') = \frac{1}{2\pi |\Sigma'|^\frac{1}{2}} \exp\left(-\frac{1}{2}g(\bm{x}|\bm{\mu}',\Sigma')\right)\,,
\end{equation}
\begin{equation}
    g(\bm{x}|\bm{\mu},\mathsf{\Sigma}) = (\bm{x}-\bm{\mu})^{\top}\mathsf{\Sigma}^{-1}(\bm{x}-\bm{\mu})\,.
\end{equation}
This para-perspective projection is differentiable and denoted as the projection module in Fig.~\ref{fig:overview}. Thanks to this analytical expression of the 2D projection we can directly evaluate the discrepancy with the projection of the predicted shape, unlike methods that rely on 2D kernel density estimation of projected point clouds~\cite{navaneet2019capnet}.

We generate a pseudo soft silhouette $\hat{s}(\bm{x})$ from Eq\onedot~\eqref{eq:gmm2d} to evaluate the consistency of the projected 2D Gaussian mixture with the ground-truth silhouette $s(\bm{x}) \in [0,1]$.  Given a random sampling of $Q$ points from the 2D probability density function $d(\bm{x})$ of Eq\onedot~\eqref{eq:gmm2d}, the probability of observing at least a point out of the $Q$ points at a pixel position $\bm{x}$ is given by
\begin{equation}
    \hat{s}(\bm{x}) = 1 - \{1 - d(\bm{x})\}^Q\,.
\end{equation}
By approximating the silhouette generated from the probability density function by this $\hat{s}(\bm{x})$, we can define an L2 loss
\begin{equation}
    L_\mathrm{sil}(\hat{s}(\bm{x}), s(\bm{x})) = \sum_{\bm{x}} \left\{\hat{s}(\bm{x}) - s(\bm{x})\right\}^2\,,
    \label{eq:loss_sil}
\end{equation}
as our silhouette loss.  In the experiments, we determined $Q$ using validation data. 
In training, we use 4 random viewpoints to evaluate the 2D multi-view loss.

\section{Experimental Results}\label{sec:results}

We first describe data and metrics used for our experiments and then detail quantitative evaluation on synthetic and real images. In addition, we demonstrate 3D pose alignment and automatic level-of-detail shape recovery using 3D-GMNet.

\paragraph{Data}
For quantitative evaluation, we use 3D models in ShapeNet\cite{chang2015shapenet}. Each 3D model in ShapeNet has a polygon CAD model and its volume data. For each polygon model, multi-view RGB images are rendered from random 100 viewpoints at a unit distance from the model using a tessellated icosahedron. To normalize the apparent size of the model in the rendered images, the distance is adjusted on a per-model basis as the diagonal size of the object bounding box. The virtual camera is configured as $128 \times 128$ resolution and $68^\circ$ field-of-view. We train and evaluate our network for 4 object categories, namely \textit{Chair}, \textit{Car}, \textit{Airplane} and \textit{Table}. For real images, we evaluate our method using real chair images in Pix3D Dataset\cite{pix3d}. We remove images in which the object is partly in the image or occluded by other objects for simplicity. Occlusion handling will be addressed in future work. We resize and crop images using manually annotated 2D masks. 

\iftrue
\begin{table}[b]
\small
\setlength{\tabcolsep}{3pt}
    \centering
    \subfloat[][]{
        \begin{tabular}{|l|rr|} \hline
            \rule{0pt}{10.1mm} & 
            \multicolumn{1}{c}{\footnotesize\raisebox{3.2mm}{3D}} & \multicolumn{1}{c|}{\footnotesize\raisebox{3.2mm}{3D+MV}}\\ \hline
            { CD} & 0.0866  &  \textbf{0.0842} \\ 
            { EMD} &0.0923  &  \textbf{0.0889} \\
            { IoU} &0.466  &  \textbf{0.482} \\
            \hline
        \end{tabular}
        \label{table:sub_sil}
    }
    \subfloat[][]{
    \begin{tabular}{|ccccccc|c|} \hline
            \multicolumn{1}{|c}{\scriptsize \begin{tabular}{c}3D-\\R2N2\\\cite{choy20163d}\end{tabular}
} & 
            \multicolumn{1}{c}{\scriptsize \begin{tabular}{c}PSGN\\\cite{Fan_2017_CVPR}\end{tabular}} & 
            \multicolumn{1}{c}{\scriptsize \begin{tabular}{c}3D-\\VAE-\\GAN\\\cite{3d-vae-gan}\end{tabular}} & 
            \multicolumn{1}{c}{\scriptsize \begin{tabular}{c}DRC\\\cite{Tulsiani_2017_CVPR}\end{tabular}} & 
            \multicolumn{1}{c}{\scriptsize \begin{tabular}{c}Marr\\Net\\\cite{marrnet}\end{tabular}} & 
            \multicolumn{1}{c}{\scriptsize \begin{tabular}{c}Atlas\\Net\\\cite{atlasnet}\end{tabular}} & 
            \multicolumn{1}{c|}{\scriptsize \begin{tabular}{c}Pix3D\\\cite{pix3d}\end{tabular}} & 
            \multicolumn{1}{c|}{\scriptsize Ours}\\
        \hline
        0.239 & 0.200 & 0.182 & 0.160 & 0.144 & \textcolor{blue}{0.125} & \textbf{0.119} & 0.130\\
        0.211 & 0.216 & 0.176 & 0.144 & 0.136 & \textcolor{blue}{0.128} & \textbf{0.120} & 0.129\\
        0.136 & N/A & 0.171 & \textcolor{blue}{0.265} & 0.231 & N/A & \textbf{0.287} & 0.259\\ \hline
    \end{tabular}
    \label{table:sub_real}
    }
    \vspace{5pt}
    \caption[]{\subref{table:sub_sil} Reconstruction accuracy only using the 3D Gaussian mixture loss (3D) and also with the 2D multi-view loss (3D+MV). The 2D multi-view loss increases reconstruction accuracy. \subref{table:sub_real} Accuracy of single image 3D reconstructions using real images in Pix3D dataset as reported in~\cite{pix3d}. 3D-GMNet (Ours) achieves accuracy comparable to state-of-the-art but with significantly smaller memory footprint and flexible representation.}
    \label{table:results}
\end{table}
\else
\begin{table}
  \begin{center}
    \caption{Quantitative evaluation of reconstruction accuracy only using the 3D Gaussian mixture loss (3D) and also with the 2D multi-view loss (3D+MV). The 2D multi-view loss increases reconstruction accuracy.}
    \begin{tabular}{|l|rrr|} \hline
        & \multicolumn{1}{|c}{CD} & \multicolumn{1}{c}{EMD} & \multicolumn{1}{c|}{IoU} \\ \hline
        3D & 0.0866  &  0.0923  &  0.466 \\ 
        3D + MV &0.0842  &  0.0889  &  0.482 \\
        \hline
    \end{tabular}
    \label{table:result_sil}
  \end{center}
\end{table}
\begin{table}
  \begin{center}
    \caption{Quantitative evaluation of single image 3D reconstructions using real images in Pix3D dataset\cite{pix3d}. Top 7 rows are the results reported by Sun \etal\cite{pix3d}.  The last line is the result with a data augmentation in the training ShapeNet data so that the object appears in different scales in the image as Pix3D images do.}
    \begin{tabular}{|l|rrrrrrr|rr|} \hline
        & 
            \multicolumn{1}{|c}{\begin{tabular}{c}3D-\\ R2N2 \\ \cite{choy20163d}\end{tabular}
} & 
            \multicolumn{1}{c}{\cite{Fan_2017_CVPR}} & 
            \multicolumn{1}{c}{\cite{3d-vae-gan}} & 
            \multicolumn{1}{c}{\cite{Tulsiani_2017_CVPR}} & 
            \multicolumn{1}{c}{\cite{marrnet}} & 
            \multicolumn{1}{c}{\cite{atlasnet}} & 
            \multicolumn{1}{c|}{\cite{pix3d}} & 
            \multicolumn{1}{|c}{Ours} & 
            \multicolumn{1}{c|}{Ours*}\\
        \hline
        IoU & 0.000 & 0.000 & 0.000 & 0.000 & 0.000 & 0.000 & 0.000 & 0.000 & 0.000\\
        EMD & 0.000 & 0.000 & 0.000 & 0.000 & 0.000 & 0.000 & 0.000 & 0.000 & 0.000\\
        CD & 0.000 & 0.000 & 0.000 & 0.000 & 0.000 & 0.000 & 0.000 & 0.000 & 0.000\\ \hline
    \end{tabular}
    \label{table:result_real}
  \end{center}
\end{table}
\fi

\begin{figure}[tb]
\begin{center}
\includegraphics[clip,width=\linewidth]{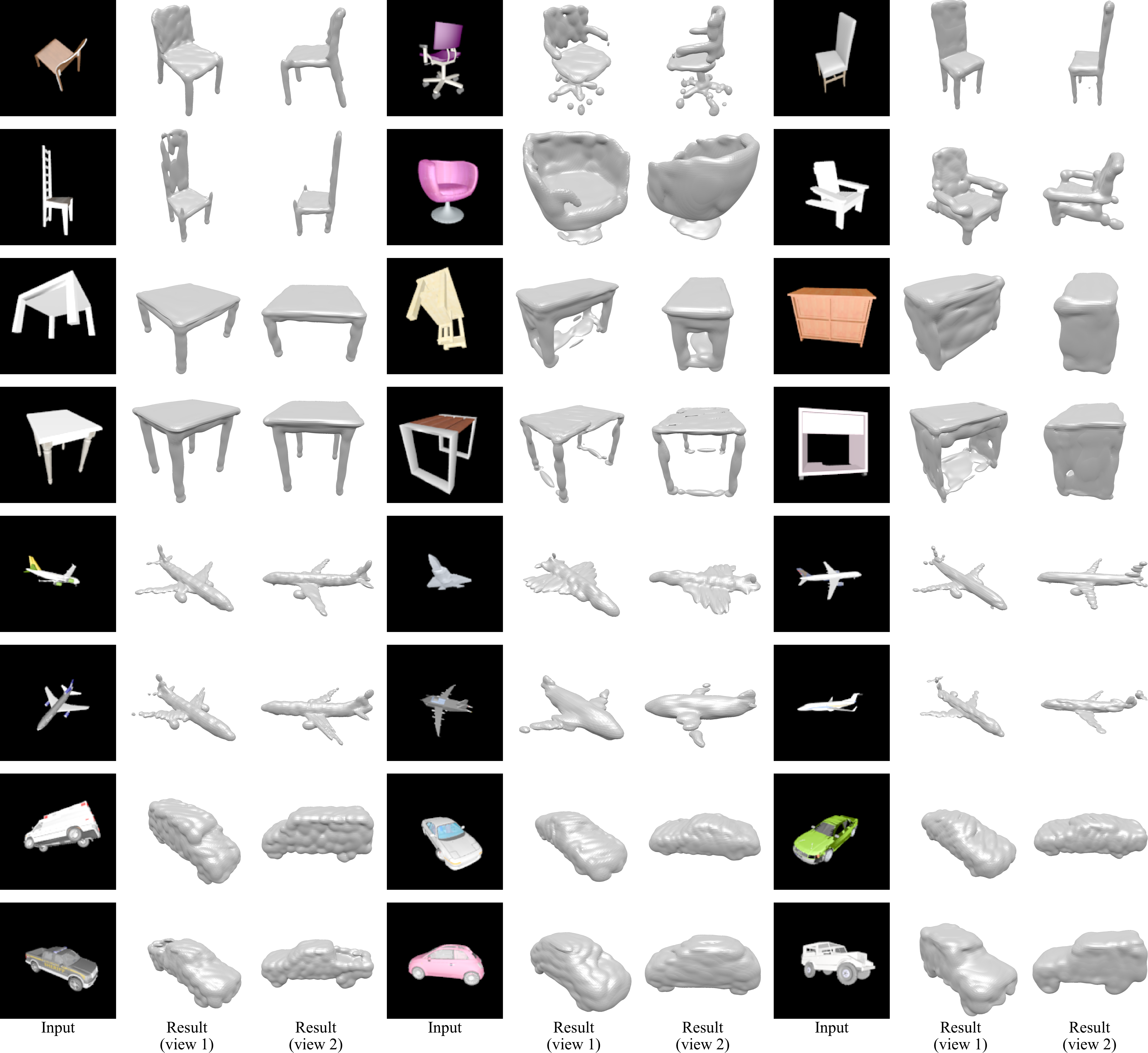}
\caption{Class-specific single-image 3D reconstruction for 4 object categories with $K=256$ Gaussian mixture shown as surface meshes from two novel views. 3D-GMNet accurately recovers complex shape including angled and thin structures.}
\label{fig:result_singleimage}    
\end{center}
\end{figure}
\begin{figure}[t]
\begin{center}
\includegraphics[clip,width=\linewidth]{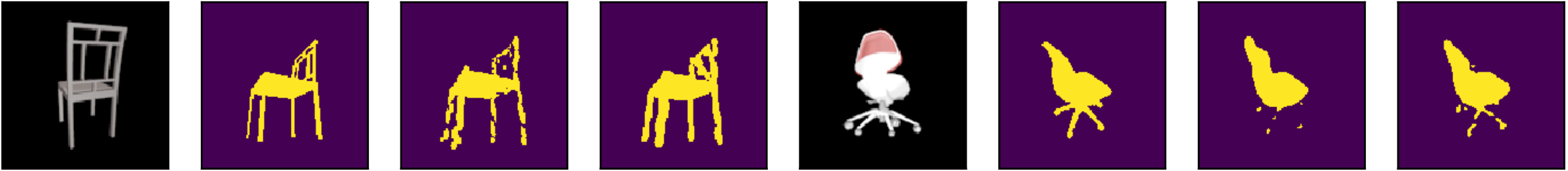}
\caption{Effectiveness of 2D multi-view loss (from left to right, input image, ground truth silhouette, silhouette of reconstruction without and with the 2D multi-view loss). The 2D multi-view loss is essential for recovering complex structures such as thin chair legs. }
\label{fig:result_sil}    
\end{center}
\end{figure}

Following \cite{pix3d}, we use three metrics for evaluation: intersection of union (IoU), earth mover's distance (EMD), and chamfer distance (CD). IoU evaluates the coverage of the estimated volume \wrt the ground truth volume, using voxelized Gaussian mixture. Higher IoU means better reconstruction results. EMD and CD evaluate geodesic and shortest distances between two surfaces via point clouds sampled on them, respectively. As described in \cite{pix3d}, we uniformly sampled points on the estimated and the ground truth surface to generate a dense point cloud, and then randomly sampled $1024$ points from the point cloud. They are scaled to fit a unit cube for normalization for EMD and CD calculation.  We used the implementation by Sun \etal\cite{pix3d}.



\vspace{-12pt}
\paragraph{Training Parameters}
We use the Adam optimizer with learning rate of $10^{-4}$. The mini batch size is set to $64$. Training loss is averaged in each mini batch. We use $80\%$ of the 3D models in ShapeNet for training, $10\%$ for validation, and the rest for testing.

\subsection{Single-Image 3D Reconstruction}

Fig.~\ref{fig:result_singleimage} shows predicted 3D models. Given the single input image 3D-GMNet estimates the object shape as a 3D Gaussian mixture, which is rendered from two novel views as a mesh model. The renderings from the novel views demonstrate qualitatively that the proposed 3D-GMNet can estimate the full 3D shape including thin, angled structures accurately.

\vspace{-12pt}
\paragraph{Contribution of 2D Multi-View Loss}
Table \ref{table:results}\subref{table:sub_sil} shows shape reconstruction accuracy using only the 3D Gaussian mixture loss (3D) and also with the 2D multi-view loss (MV). Fig.~\ref{fig:result_sil} shows silhouettes of reconstructed 3D shapes with and without the 2D multi-view loss. These results clearly show that the silhouette loss reduces 3D shape reconstruction errors and enables recovery of complex geometric structures.


\vspace{-12pt}
\paragraph{Number of Gaussian Components}
\begin{figure}
    \centering
    \includegraphics[width=\linewidth]{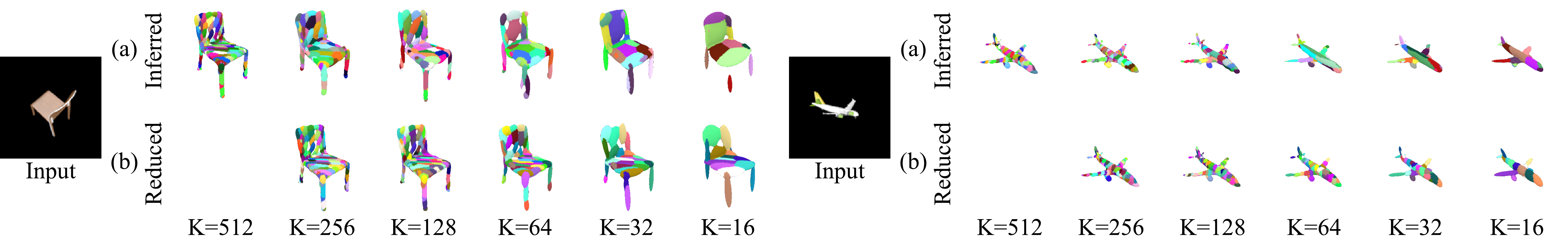}
    \caption{(a) 3D shapes estimated with different numbers of Gaussian components. Different color indicates a distinct Gaussian component. (b) Controlled level-of-detail reconstruction. Results of Gaussian mixture reduction from $K=512$ to $K=16,32,\dots,256$. The results show that we can control the level-of-detail of the reconstruction without altering the network while maintaining accuracy.}
    \label{fig:result_reduction_comp-crop}
\end{figure}

\begin{figure}[tb]
    \centering
    \subfloat[][]{
        \includegraphics[clip,height=60pt]{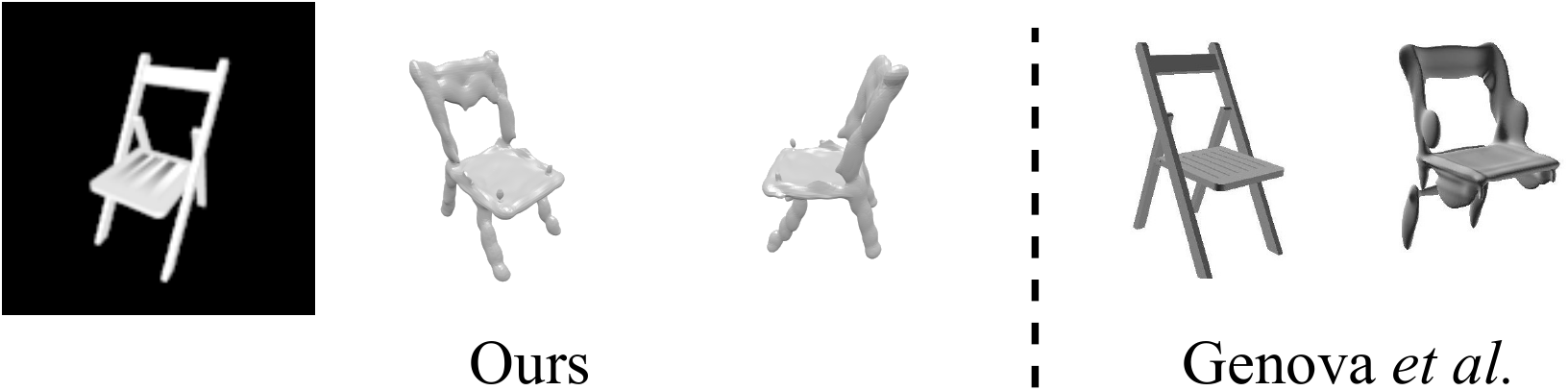}
        \label{fig:sub_genova}
    }
    \\
    \subfloat[][]{
        \raisebox{5pt}{\includegraphics[clip,width=\linewidth]{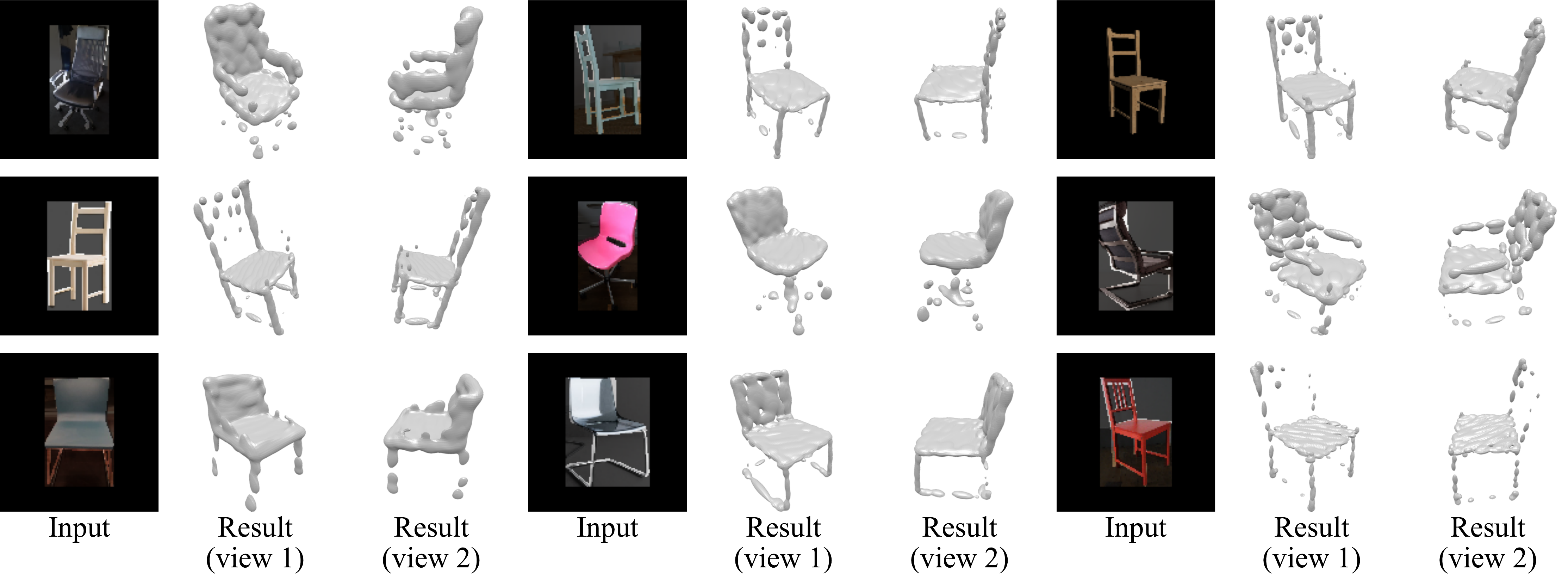}}
        \label{fig:sub_real}
    }
    \vspace{5pt}
    \caption[]{\subref{fig:sub_genova} Left : Input image and output isosurface of our method. Right : Input mesh and reconstruction result reported in Genova \etal\cite{Genova_2019_ICCV}. \subref{fig:sub_real} Reconstructions from real images in the Pix3D dataset\cite{pix3d} shown as surface mesh models rendered from two novel views.}
    \label{fig:result_comp_real}    
\end{figure}

Fig.~\ref{fig:result_reduction_comp-crop}(a) shows recovery 3D Gaussian-mixture shape models using 3D-GMNet with different numbers of mixture components $K$. We can observe that though reconstructions with higher $K$ results in a detailed reconstruction, those with lower $K$ also approximates the 3D shape accurately.

\vspace{-12pt}
\paragraph{Comparison to Genova \etal}
Fig.~\ref{fig:result_comp_real}\subref{fig:sub_genova} shows qualitative comparison of reconstructed 3D shape of a chair from a single using 3D-GMNet and from its mesh model as shown in Genova \etal . Our reconstruction is not limited to axis-aligned Gaussians, which results in superior reconstruction even from a single image.

\vspace{-12pt}
\paragraph{3D Reconstruction from Real Images}
Fig.~\ref{fig:result_comp_real}\subref{fig:sub_real} and Table \ref{table:results}\subref{table:sub_real} show single-image 3D reconstruction results with real images in the Pix3D dataset\cite{pix3d}. Note that the training scheme (trained for a single category or multiple categories, in object-centered manner or viewer-centered manner) differs for each method. For this, the quantitative comparison is not necessarily fair. In addition to the several advantages of our shape representation, the reconstruction accuracy of 3D-GMNet is comparable to the state-of-the-art methods.


\subsection{3D Pose Estimation}\label{sec:alignment}

\begin{figure}[tb]
\begin{center}
\includegraphics[clip,width=\linewidth]{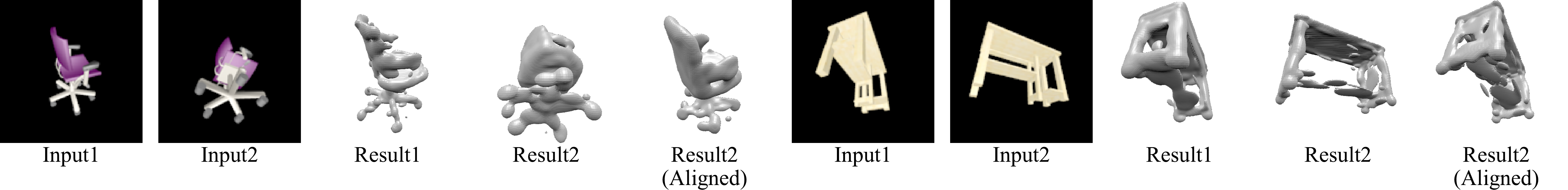}
\caption{3D pose estimation results. 3D-GMNet enables estimation of relative camera pose between input images (input2 to input1).}
\label{fig:result_pose_alignment}    
\end{center}
\end{figure}

3D-GMNet recovers the shape in the local camera coordinate system of the input image. Given two images of a single object from different viewpoints, 3D-GMNet can recover the 3D object shape in two different coordinate systems, which means that we can estimate the relative pose of the cameras by aligning the estimated 3D shapes. The key challenge for achieving this pose estimation is the view-dependent assignment of Gaussian components in the recovered 3D Gaussian mixture. We solve this by aligning the covariance matrices of Gaussian mixtures from different viewpoints. In the supplemental material, we show that this can be computed analytically. Fig.~\ref{fig:result_pose_alignment} shows the alignment results. The results show that our method can provide reasonable pose alignments without explicit point cloud generation.


\subsection{Controlled Level-of-Detail Reconstruction}
Fig.~\ref{fig:result_reduction_comp-crop}(b) demonstrates controlling the level-of-detail of shape reconstruction by automatically varying the number of components of the Gaussian mixture shape model. Given an input image (the leftmost column), our network with $K=512$ infers a Gaussian mixture representation of its 3D shape (the second column). By applying  Gaussian mixture reduction\cite{runnalls2007kullback}, we can obtain different level-of-details of the underlying 3D shape as shown in the second and the fourth rows. When compared with the 3D shapes estimated by 3D-GMNet originally trained with the corresponding number of components (the first and the third rows), the controlled level-of-detail reconstruction yield similar accuracy.

\section{Conclusion}
We proposed 3D-GMNet for recovering the 3D shape of an object from its single-view observation as a Gaussian mixture. We introduced a 3D Gaussian mixture loss and a 2D multi-view loss to accurate reconstruct the 3D shape from a single image. Experimental results show that our 3D-GMNet successfully estimates the object 3D shape as a compact Gaussian mixture that can be sampled and viewed as conventional geometry representations including point cloud, volume, and mesh model. Extensive experimental validation showed that the method can recover 3D shape accurately even with a lower number of components, while maintaining comparable performance with state-of-the-art methods, but with the additional benefits of this unique shape representation including pose estimation and level-of-detail control.

\paragraph*{Acknowledgement}
This work was in part supported by JSPS KAKENHI 17K20143 and JST PRESTO Grant Number JPMJPR1858.

\bibliography{egbib}

\end{document}